%% file: arxiv.tex
\documentclass[runningheads]{llncs}
\usepackage{graphicx}
\usepackage{comment}
\usepackage{amsmath,amssymb} 
\usepackage{color}
\usepackage{subfigure}
\usepackage{multirow}
\usepackage{float}

\newcommand{\tabincell}[2]{\begin{tabular}{@{}#1@{}}#2\end{tabular}}

\usepackage[pagebackref=false,breaklinks=true,letterpaper=true,colorlinks,bookmarks=false]{hyperref}

\begin{document}
\pagestyle{headings}
\mainmatter
\def\ECCVSubNumber{****}  

\title{Spatially Adaptive Inference with Stochastic Feature Sampling and Interpolation} 

\titlerunning{ }
%
\author{Zhenda Xie$^{1,2\dag}$\thanks{Equal contribution. \dag This work is done when Zhenda Xie and Xizhou Zhu are interns at Microsoft Research Asia. \ddag Corresponding author. } \and
Zheng Zhang$^{2\star\ddag}$\and
Xizhou Zhu$^{2,3\star\dag}$ \and
Gao Huang$^{4}$ \and
Stephen Lin$^{2}$}
\authorrunning{Zhenda Xie, Zheng Zhang, Xizhou Zhu, Gao Huang, Stephen Lin}
%
\institute{Tsinghua University \\ 
\email{xzd18@mails.tsinghua.edu.cn} \and
Microsoft Research Asia \\ 
\email{\{zhez,stevelin\}@microsoft.com} \and
University of Science and Technology of China \\
\email{ezra0408@mail.ustc.edu.cn} \and
Tsinghua University, Department of Automation \\
\email{gaohuang@tsinghua.edu.cn}}

\maketitle

\begin{abstract}
In the feature maps of CNNs, there commonly exists considerable spatial redundancy that leads to much repetitive processing. Towards reducing this superfluous computation, we propose to compute features only at sparsely sampled locations, which are probabilistically chosen according to activation responses, and then densely reconstruct the feature map with an efficient interpolation procedure. With this sampling-interpolation scheme, our network avoids expending computation on spatial locations that can be effectively interpolated, while being robust to activation prediction errors through broadly distributed sampling. A technical challenge of this sampling-based approach is that the binary decision variables for representing discrete sampling locations are non-differentiable, making them incompatible with backpropagation. To circumvent this issue, we make use of a reparameterization trick based on the Gumbel-Softmax distribution, with which backpropagation can iterate these variables towards binary values. The presented network is experimentally shown to save substantial computation while maintaining accuracy over a variety of computer vision tasks. Code is available in \url{https://github.com/zdaxie/SpatiallyAdaptiveInference-Detection}.

\end{abstract}

\section{Introduction}
\vspace{-0.5em}
On many computer vision tasks, significant improvements in accuracy have been achieved through increasing model capacity in convolutional neural networks (CNNs)~\cite{he2016deep,szegedy2015going}. These gains, however, come at a cost of greater processing that can 
hinder deployment on resource-limited devices. Towards greater practicality of deep models, much attention has been focused on reducing CNN computation.

\begin{figure}[tbp]
  \centering
  \includegraphics[width=1.00\textwidth]{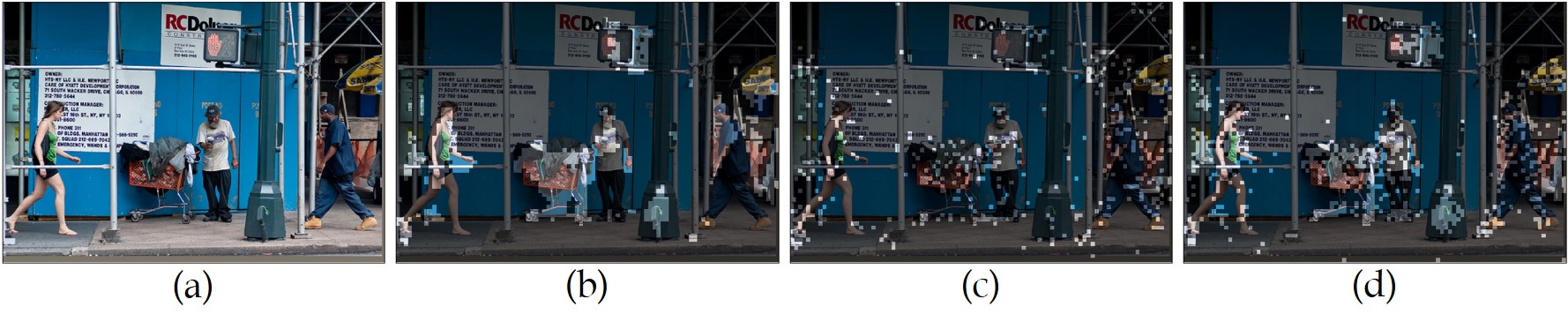}
  \vspace{-2em}
  \caption{\textbf{(a)} Input image. \textbf{(b)} Deterministic sampling for efficient inference. \textbf{(c)} Content-aware stochastic sampling by our method (without \textit{Grid Prior} for better viewing), which yields the same detection accuracy as (b) but with less overall computation. \textbf{(d)} Content-aware stochastic sampling with the same overall computation as (b) but with better detection accuracy. }
  \label{fig:teaser}
  \vspace{-1em}
\end{figure}

A common approach to this problem is to prune weights and neurons that are not needed to maintain the network’s performance~\cite{lecun1989optimal,han2015learning,guo2016dynamic,wen2016learning,he2019filter,li2017pruning,molchanov2019importance,yu2018nisp}. Orthogonal to these architectural changes are methods that eliminate computation at inference time conditioned on the input. These techniques are typically based on feature map sparsity, where the locations of zero-valued activations are predicted so that the computation at those positions can be skipped~\cite{dong2017more,ren2018sbnet,cao2019seernet}. As illustrated in Fig.~\ref{fig:teaser}(b), this approach deterministically samples predicted foreground areas while avoiding computational expenditure on the background.

In this paper, we seek a more efficient allocation of computation over a feature map that takes advantage of its spatial redundancy\footnote{We note that CNNs commonly capitalize on spatial redundancy by downsampling input image resolutions or employing strides in convolution.}. Rather than exhausting all the computation on areas estimated to have the most significant activation, our approach is to treat the predicted activation map as a probability field, stochastically sample a sparse set of locations based on their probabilities, and then interpolate the features at these samples to reconstruct the rest of the feature map. This strategy avoids wasting computation at locations where the features can simply be interpolated, and it allows feature computation to expand into presumed low-activation regions, which can compensate for errors in activation prediction. With the generated sampling distributions shown in Fig.~\ref{fig:teaser}(c) and (d), this sampling-interpolation approach can reduce the computation needed to match the accuracy of feature map sparsity methods, or alternatively increase accuracy through more comprehensive sampling of the feature map while maintaining the same level of overall computation.

To identify sparse points for interpolation, our network trains a content-aware stochastic sampling module that produces a binary sampling mask over the activations. Due to the inability to backpropagate through networks containing binary variables, we employ a reparameterization trick where the non-differentiable mask values are replaced by differentiable samples from a Gumbel-Softmax distribution, which can be smoothly annealed into binary values during training~\cite{jang2017categorical}. The module learns to spatially adjust the sampling density according to predicted activations, and the interpolation is performed with a kernel whose parameters are jointly learned with the sampling module. To aid in interpolation of areas far from the content-aware samples, very sparse uniform samples over the feature map are added before interpolation, which we refer to as a grid prior.

With this content-based sampling-interpolation approach, our network obtains appreciable reductions in computation without much loss in accuracy on COCO object detection, Cityscapes semantic segmentation, and ImageNet classification. An extensive ablation study validates the sampling and interpolation components of our algorithm, and comparisons to related techniques show that the proposed method provides a superior FLOPs-accuracy tradeoff for object detection and semantic segmentation, and comparable performance for image classification.

\vspace{-0.7em}
\section{Related work}
\vspace{-0.5em}

In this section, we briefly review related approaches for reducing computation in convolutional neural networks.

\noindent\textbf{Model pruning}
A widely investigated approach for improving network efficiency is to remove connections or filters that are unimportant for achieving high performance. The importance of these network elements has been approximated in various ways, including by connection weight magnitudes~\cite{han2015learning,guo2016dynamic}, filter norms~\cite{li2017pruning,wen2016learning,he2018soft}, and filter redundancy within a layer~\cite{he2019filter}. To reflect network sensitivity to these elements, importance has also been measured based on their effects on the loss~\cite{lecun1989optimal,molchanov2019importance} or the reconstruction error of the final response layer~\cite{yu2018nisp} when removing them. Alternatively, sparsity learning techniques identify what to prune in conjunction with network training, through constraints that zero out some filters~\cite{wen2016learning}, cause some filters to become redundant and removable~\cite{ding2019centripedal}, scale some filter or block outputs to zero~\cite{lin2019towards}, or sparsify batch normalization scaling factors~\cite{liu2017learning,ye2018rethinking}. Model pruning techniques as well as other architecture-based acceleration schemes, such as low-rank factorizations of convolutional filters~\cite{jaderberg2014speeding} and knowledge distillation of networks~\cite{hinton2015distilling}, are orthogonal to our approach and could potentially be employed in a complementary manner.

\noindent\textbf{Early stopping}
Rather than prune network elements, early stopping techniques reduce computation by skipping the processing at later stages whenever it is deemed to be unnecessary. In~\cite{figurnov2017spatially}, an adaptive number of ResNet layers are skipped within a residual block for unimportant regions in object classification. The skipping mechanism is controlled by halting scores predicted at branches to the output of each residual unit. In~\cite{li2017not}, a deep model for semantic segmentation is turned into a cascade of sub-models where earlier sub-models handle easy regions and harder cases are progressively fed forward to the next sub-model for further processing. In~\cite{kuen2018stochastic}, various predefined downsampling configurations are randomly used during training, and the appropriate configuration is applied according to the computation budget during inference. Like our method, these techniques spatially adapt the processing to the input content. However, they process all spatial positions at least to some degree, which limits the achievable computational savings.

\noindent\textbf{Activation sparsity}
The activations of rectified linear units (ReLUs) are commonly sparse. This property has been exploited for network acceleration by excluding the zero values from subsequent convolutions~\cite{shi2017speeding,parashar2017scnn}. This approach has been extended by estimating the activation sparsity and skipping the computation for predicted insignificant activations. The sparsity has been predicted from prior knowledge of road and sidewalk locations in autonomous driving applications~\cite{ren2018sbnet}, from model-predicted foreground masks at low resolution~\cite{ren2018sbnet}, from a small auxiliary layer that supplements each convolutional layer~\cite{dong2017more}, and from a highly quantized version of the convolutional layer~\cite{cao2019seernet}.
Our work instead reconstructs activation maps by interpolation from a sparse set of samples selected in a content-aware fashion, thus avoiding computation at locations where features can be easily reconstructed. Moreover, our probabilistic sampling distributes computation among feature map locations with varying levels of predicted activation, providing greater robustness to activation prediction errors.

\noindent\textbf{Sparse sampling}
To reduce processing cost, PerforatedCNNs compute only sparse samples of a convolutional layer's outputs and interpolate the remaining values~\cite{figurnov2016perforatedcnns}. The sampling follows a predefined pattern, and the interpolation is done by nearest neighbors. Our method also takes a sparse sampling and interpolation approach, but in contrast to the input-independent sampling and generic interpolation of PerforatedCNNs, the sampling in our network is adaptively determined from the input such that the sampling density reflects predicted activation values, and the interpolation parameters are learned. As shown later in the experiments, this approach allows for much greater sparsity in the sampling. In~\cite{mazzini2019spatial}, high-resolution predictions are generated from low-resolution results. Instead, our method is used for features rather than final outputs.

\noindent\textbf{Gumbel-based selection}
Random selection based on the Gumbel distribution has been used in making discrete decisions for network acceleration. The Gumbel-Softmax trick was utilized in adaptively choosing network layers to apply on an input image~\cite{veit2017convolutional} and in selecting channels or layers to skip~\cite{herrmann2019end-to-end}. In contrast to these techniques which determine computation based on image-level semantics for image classification, our sampling is driven by the spatial organization of features and is geared towards accurately reconstructing positional content. As a result, our method is well-suited to spatial understanding tasks such as object detection and semantic segmentation.

\vspace{-0.7em}
\section{Methodology}
\vspace{-0.5em}
\input{tex/methodology.tex}

\vspace{-0.7em}
\section{Experiments}
\vspace{-0.5em}
\input{tex/experiments.tex}

\vspace{-0.7em}
\section*{Acknowledgment}
\vspace{-0.5em}
Gao Huang is supported in part by Beijing Academy of Artificial Intelligence (BAAI) under grant BAAI2019QN0106. In addition, we would like to thank Jifeng Dai for his early contribution to this work during his work at Microsoft Research Asia. Jifeng later turned to other exciting projects.

\appendix
\renewcommand{\thesection}{A\arabic{section}}   
\vspace{-0.7em}
\section{Experimental Settings}
\vspace{-0.5em}
\noindent \textbf{Object detection}
All models are trained on the 118k images of the COCO 2017~\cite{lin2014coco} train set, and evaluated on the 5k images of the COCO 2017 validation set. The standard mean average precision (mAP) is used to measure accuracy. The baseline model is based on Faster R-CNN with Feature Pyramid Network (FPN)~\cite{lin2016feature} and deformable convolution~\cite{dai2017deformable}. ImageNet~\cite{deng2009imagenet} pre-trained ResNet-101~\cite{he2016deep} is chosen as the default backbone model. Following \cite{zhu2019deformable}, all the $3\times3$ convolutions from the conv3 to conv5 stages are replaced by deformable convolutions. 

The implementation and hyper-parameters are based on mmdetection~\cite{mmdetection2018}. Anchors with 5 scales and 3 aspect ratios are used. 2k and 1k region proposals are generated with a non-maximum suppression threshold of 0.7 at training and inference, respectively. The network is trained for 12 epochs on 8 GPUs with 1 image per GPU. In SGD training, the learning rate is initialized to 0.01 and is divided by 10 at the 8th and 11th epochs. The weight decay and the momentum parameters are set to $0.0001$ and $0.9$, respectively. 

\noindent\textbf{Semantic segmentation}
All models are trained on the $2975$ finely annotated images of the Cityscapes~\cite{cordts2016cityscapes} train set and evaluated on the $500$ images of the validation set. Accuracy is measured by the standard mean IoU. The baseline model is ResNet-50 based dilated FCN~\cite{long2015fully} with deformable convolution layers~\cite{dai2017deformable}. Following \cite{zhu2019deformable}, we use deformable convolutions to replace all $3\times3$ convolutions from the conv3 to conv5 stages; the strides of the conv4 and conv5 stages are set to 1, and the dilations of all $3\times3$ convolutions in the conv4 and conv5 stages are set to 2 and 4, respectively.

The implementation and hyper-parameters are based on the open-source implementation of TorchCV~\cite{you2019torchcv}. The networks are trained on 4 GPUs with 2 images per GPU for 60k iterations. The SGD optimizer with the poly learning rate policy is employed. The initial learning rate is 0.01 and the decay exponent is 0.9. The weight decay and momentum are set to 0.0001 and 0.9, respectively. In the training stage, the data is augmented with random scaling (from $0.5$ to $2.0$), random cropping and random horizontal flipping. Synchronized Batch Normalization~\cite{peng2018megdet} with learnable weights is placed after every newly added layer. 

\vspace{-1em}
\section{Numerical Results}
\vspace{-0.5em}
\input{table/object_res}
\noindent\textbf{Object detection}
Numerical results of Fig.~\ref{fig:det_seg_results} (a) in the main paper are shown in Table~\ref{table.object_res}. For the ResNet-50 model, we resize the sides of input images to 1000. For uniform sampling, we resize the shorter sides of input images to \{1000, 800, 600, 500\} to generate results under different FLOPs. For our method and the two deterministic methods (deterministic Gumbel-Softmax and ReLU), we obtain results by adopting different sparse loss weights, i.e. \{0.2, 0.1, 0.05, 0.02\}.

\input{table/seg_res}
\noindent\textbf{Semantic segmentation}
The numerical results of Fig.~\ref{fig:det_seg_results} (b) in the main paper are shown in Table~\ref{table.seg_res}. For uniform sampling, we resize the shorter sides of input images to \{1024, 896, 736, 512\} to obtain results under different FLOPs. For our method and deterministic Gumbel-Softmax, we generate results by adopting different sparse loss weights, i.e. \{0.3, 0.2, 0.1, 0.05\}. 

\input{table/inference_stability}
\vspace{-1em}
\section{Inference Stability}
\vspace{-0.5em}
Our stochastic Gumbel-Softmax sampling method has randomness in the inference phase, so the results may differ with each test. In this section, we evaluate the inference stability of our model (ResNet-101) on object detection (COCO2017 validation). We evaluate multiple models, which were trained by different loss weights. For each model, we test five times and report the mean and standard deviation of the mAP and FLOPs. The results are shown in Table~\ref{table.inf_stab}. For all models, the variance of mAP and FLOPs is small, indicating that our method has strong stability in the inference phase. We speculate that this stability may arise from the grid prior. Therefore, we further evaluate the testing stability without the grid prior and found that the variance of mAP over multiple tests increased from about 0.02 to 0.09. 

\vspace{-1em}
\section{Analysis on Final Temperature}
\vspace{-0.5em}
Intuitively, the decay factor should be related to the training dynamics, rather than to the tasks. Ideally, the decay factor should not be too large or too small. If it is too small, the temperature will quickly drop to near zero, and then the mask will degenerate to binary, resulting in no gradient and insufficient training. On the other hand, if the factor is too large, the temperature will still be high at the end of training, and the mask is not binary in this case, which will result in inconsistency between inference and training. In principle, we want the temperature to be close to 0 at the end of training.

In practice, we tried different decay factors on COCO, so that the final temperature at the end of training is 0.03, 0.01 or 0.005. We found the difference in performance to be small (less than 0.2 mAP and 3 GFLOPs for different models), showing that the final temperature has little effect on the results within a reasonable interval. In addition, we did not tune the final temperature on other tasks, but directly adopted 0.01 and found it to work well. This may partly indicate that the decay factor is insensitive to different tasks.

\vspace{-1em}
\section{Performance on ResNeXt}
\vspace{-0.5em}
\input{figures/resnext_det.tex}
To examine the compatibility of our method with other network architectures, we integrate it with ResNeXt~\cite{xie2017aggregated} and evaluate the performance on object detection. 
ImageNet pre-trained ResNeXt-101($32\times4d$) with deformable convolution is chosen as the backbone model. We use the same hyper-parameters and training setting as in Sec.~4.1 of the main paper. For uniform sampling models, we resize the shorter sides of input images to \{1000, 800, 600, 500\}. For our method, we use different loss weights \{0.2, 0.1, 0.05, 0.02\} to draw the speed-accuracy curve. The results are shown in Fig.~\ref{fig:resnext_det} and the numerical results are shown in Table~\ref{table.resnext_res}. Our method is found to outperform uniform sampling by a large margin.
\input{table/resnext_res}

\vspace{-1em}
\section{Sparsity of different blocks}
\vspace{-0.5em}
We evaluate the average sparsity of different blocks on the COCO2017 validation set for models trained under different loss weights. Results are shown in Fig.~\ref{fig:sparsity_diff_layers}. For all the models, we observed that deeper blocks are more sparse, and this phenomenon is more pronounced in models trained with small loss weights. For example, in the model trained with a loss weight of 0.02, the sparsity of conv1 is close to 0 from Res1-Block1 to Res2-Block4. One possible reason is that shallow blocks primarily compute local visual features (such as edges and textures), so their spatial redundancy is small, while deeper blocks are more likely to capture semantic features which are more redundant. Another observation is that the sparsity of conv2 and conv3 are always greater than the sparsity of conv1. The reason is unclear but we suspect that this phenomenon may be related to the receptive field of operators.
\input{figures/sparsity_diff_layers.tex}

\clearpage
%
%
\bibliographystyle{splncs04}
\bibliography{egbib}
\end{document}

%% file: tex/methodology.tex
In this section, we first present a general introduction of the stochastic sampling-interpolation network and then describe the stochastic sampling module and interpolation module in detail. Next, we introduce the grid prior which is found to be helpful for interpolation. Finally, we illustrate how to integrate the sampling and interpolation network modules into residual blocks. 

\vspace{-1em}
\subsection{Stochastic Sampling-Interpolation Network}
\vspace{-0.5em}
\input{figures/Pipeline-and-Sample-Comparison}
Convolutions in neural networks typically generate an output feature map $Y\in\mathbb{R}^{C_{out} \times H \times W}$ pixel by pixel from an input feature map $X\in\mathbb{R}^{C_{in} \times H \times W}$:
\vspace{-0.1em}
\begin{equation}
Y(p) = \sum_{p' \in R_k} W_c(p')X(p+p'), p \in \Omega,
\label{equ:conv}
\end{equation}
\vspace{-0.1em}
\noindent where $H$ and $W$ represent the height and width of the feature map, $C_{in}$ and $C_{out}$ denote the input and output feature dimensions, $\Omega=\{(i,j) |i\leq W, j\leq H, i,j\in\mathbb{Z}^+\}$ represents the spatial domain of the feature map, $R_k$ indicates the support region of kernel offsets with kernel size $k$ (e.g., for a $3\times3$ convolution, $R_k=\{(-1,-1), (-1, 0),...,(1,1)\}$ and $k=3$), and $W_c\in\mathbb{R}^{C_{in} \times C_{out} \times k \times k}$ denotes convolution weights.

Spatial redundancy commonly exists in feature maps, such that features at certain points can be approximated by interpolating the features from surrounding positions. Therefore, exhaustive computation across the entire space is not required. Our method takes advantage of this using a content-aware stochastic sampling module and a trainable interpolation module. 

A basic illustration of our method is shown in Fig.~\ref{fig:pipline-and-sample-comparison}(a). The sampling module generates a mask $\mathcal{M} \in \mathbb{R}^{H \times W}$. In the inference phase, $\mathcal{M}$ is binary and it is calculated first before computing $Y$. Points masked as $1$ in $\mathcal{M}$ are sampled, and convolution is applied only on these points, resulting in a sparse feature map $Y_s$, which is calculated as:
\vspace{-0.1em}
\begin{equation}
Y_s(p) =\left\{
\begin{array}{rcl}
0       &      & {\mathcal{M}(p) = 0}\\
Y(p)       &      & {\mathcal{M}(p) = 1}.
\end{array} \right.
\end{equation}
\vspace{-0.1em}
Then, the features of unsampled points are constructed by the interpolation module $C$. Together with the features of sampled points, they constitute the reconstructed output feature $Y^*$:  
\vspace{-0.1em}
\begin{equation}
Y^*(p) =\left\{
\begin{array}{rcl}
C(Y_s)(p)       &      & {\mathcal{M}(p) = 0}\\
Y_s(p)       &      & {\mathcal{M}(p) = 1}.
\end{array} \right.
\label{equ:interpolation}
\end{equation}
\vspace{-0.1em}
Since the cost of calculating $\mathcal{M}$ is less than that of $Y$ and $C$, if $\mathcal{M}$ is sparse, then the computation cost can be reduced. In our experiments, the sparsity of $\mathcal{M}$ can be greater than $70\%$ on average.

A technical challenge of this approach is that the binary sampling mask $\mathcal{M}$ is non-differentiable, making this sampling module incompatible with backpropagation in the training stage. To circumvent this issue, a mask $\mathcal{M}$ that gradually changes from soft to hard during training is used. Therefore, the mask can undergo optimization from the beginning of training and then becomes roughly consistent to a hard mask used in the inference stage by the end of training. During training, with this soft mask, the output $Y_s$ at each point $p$ is calculated as follows:
\vspace{-0.1em}
\begin{equation}
    Y_s(p) = \mathcal{M}(p)\odot Y(p),
\end{equation}
\vspace{-0.1em}
and the full output feature $Y^*$ is calculated by:
\vspace{-0.1em}
\begin{equation}
    Y^*(p) = (1-\mathcal{M}(p))\odot C(Y_s)(p) + \mathcal{M}(p)\odot Y_s(p),
\label{equ:soft-output}
\end{equation}
\vspace{-0.1em}
\noindent where $\odot$ denotes broadcast multiplication.

\vspace{-1em}
\subsection{Stochastic Sampling Module}
\label{sec:stochastic_sampling_module}
\vspace{-0.5em}
In previous works~\cite{ren2018sbnet,dong2017more,cao2019seernet}, sampling is usually done in a deterministic manner, where points with confidence greater than a certain threshold are sampled, as shown in Fig.~\ref{fig:pipline-and-sample-comparison}(b)(Left). But in our stochastic sampling, a higher confidence only indicates a higher probability of the point being sampled, as shown in Fig.~\ref{fig:pipline-and-sample-comparison}(b)(Right). Due to the spatial redundancy over the feature map, adjacent points may have similar features and confidences, so deterministic sampling typically samples or not samples adjacent points together. However, stochastic sampling can sample a portion of the points, and the features of the other unsampled points can be obtained from the interpolation module. Therefore, sparser sampling can be achieved while maintaining relatively accurate feature maps.

Our sampling is based on the two-class Gumbel-Softmax distribution, which was first introduced in reinforcement learning~\cite{jang2017categorical} to simulate stochastic discrete sampling. Thus, the mask $\mathcal{M}$ is defined as:
\vspace{-0.1em}
\begin{equation}
\mathcal{M}(p) = \frac{exp((-log(\pi_1(p)) + g_1(p))/\tau)}{\sum_{j\in \{0,1\}} exp((-log(\pi_j(p)) + g_j(p))/\tau)},
\label{equ:gumbel_sigmoid}
\end{equation}
\vspace{-0.1em}
\noindent where $\pi$ denotes confidence map generated from a two-class Softmax activation. $g$ represents noise sampled from a standard Gumbel distribution, and it provides the randomness of Gumbel-Softmax. If the noise $g$ is eliminated, Gumbel-Softmax degenerates into a deterministic function that is approximately equal to the Softmax function with a temperature term. $\tau$ is a temperature parameter. When $\tau$ approaches $0$, $\mathcal{M}(p)$ becomes approximately binary.

In our implementation, $\pi$ is generated by a $3\times3$ convolutional layer with a two-class Softmax activation, and $\tau$ is exponentially decreased over iterations according to $\tau=\alpha^{iter}\tau_0$, where $\alpha$ is the decay factor, $iter$ is the number of iterations, and $\tau_0$ is the initial temperature. In our experiments, we set $\tau_0=1$. Therefore, at the beginning of training, the mask is soft, allowing the sampling module to be trained. At the end of training, $\tau$ becomes close to $0$, so the mask generated by the sampling module is nearly binary, as desired for our discrete inference. To encourage the network to produce sparse sampling masks, the sparse loss is introduced on the confidence map $\pi_1$ for all layers during training:
\vspace{-0.1em}
\begin{equation}
    L_{sparse} = \sum_{l}{\|\pi^{l}_{1}\|_{1}}
\end{equation}
\vspace{-0.1em}
\noindent where $\pi^{l}_{1}$ indicates $l$-th layer's confidence map, and $\|\cdot\|_{1}$ indicates L1-norm. Different levels of sparsity are achieved by adjusting the weight of the sparse loss. Therefore, the training objective is:
\vspace{-0.1em}
\begin{equation}
    L = L_{task} + \gamma L_{sparse}.
\end{equation}
\vspace{-0.1em}
\noindent where $L_{task}$ is the task specific objective and $\gamma$ is the sparse loss weight.

\vspace{-1em}
\subsection{Interpolation Module}
\label{sec:interpolation_module}
\vspace{-0.5em}
In previous works~\cite{dong2017more,figurnov2017spatially}, the features of unsampled points are obtained by reusing the previous features at the corresponding points~\cite{figurnov2017spatially}:
\vspace{-0.1em}
\begin{equation}
Y^*(p)=\mathcal{M}(p) \odot Y_s(p) + (1-\mathcal{M}(p)) \odot X(p)
\label{equ:reuse_feature}
\end{equation}
\vspace{-0.1em}
or just by setting them to zero~\cite{dong2017more}:
\vspace{-0.1em}
\begin{equation}
\small
Y^*(p)=\mathcal{M}(p) \odot Y_s(p) + (1-\mathcal{M}(p)) \odot \boldsymbol{0} = \mathcal{M}(p) \odot Y_s(p).
\label{equ:fill_zero}
\end{equation}
\vspace{-0.1em}
However, these approaches ignore the spatial redundancy of feature maps, which could be used to obtain the features of unsampled points, leading to a more accurate feature map.

Our method capitalizes on this spatial redundancy to generate relatively accurate feature maps. This is done by interpolating the features of unsampled points from those of sampled points as indicated in Eq.~(\ref{equ:interpolation}). The interpolation is formulated as: 
\vspace{-0.1em}
\begin{equation}
C(Y_s)(p) = \frac{\sum_{s_i} W_I(p,s_i)Y_s(s_i)}{\sum_{s_i} W_I(p, s_i) + \epsilon}, s_i\in \Omega,
\label{equ:interpolation_detail}
\end{equation}
\vspace{-0.1em}
where $W_I(p,s_i) \geq 0$ represents interpolation weights, $\Omega$ represents the spatial domain like in Eq.~(\ref{equ:conv}), and $\epsilon$ is a small constant which is set to 10$^{-5}$. In this formula, the features of the unsampled points are represented by a weighted average of the features at sampled points. However, interpolation by considering all sampled points is costly, even if $Y_s$ is sparse in the inference phase. Fortunately, since neighboring points commonly have stronger feature correlation, we can alleviate this problem by computing the interpolation only using samples that lie within a window centered on the given unsampled point. Specifically, we restrict $s$ to a window of radius $r$ centered on the unsampled point $p$, defined as $\mathcal{R}_s^r(p)=\{s_i|s_i \in \Omega, ||s_i-p||_{\infty} \leq r\}$. Thus, the windowed interpolation module is formulated as:
\vspace{-0.1em}
\begin{equation}
C(Y_s)(p) = \frac{\sum_{s_i} W_I(p,s_i)Y_s(s_i)}{\sum_{s_i} W_I(p, s_i)}, s_i\in \mathcal{R}_s^r(p).
\label{equ:windowed_interpolation_detail}
\end{equation}
\vspace{-0.1em}

The best formulation of $W_I$ in irregular spatial sampling is an open problem. We explored different design choices:

\noindent\textbf{RBF Kernel} The radial basis function (RBF) kernel is a commonly used interpolation function, defined as:
\vspace{-0.1em}
\begin{equation}
    W_I(p_1, p_2)=exp(-\lambda^2||p_1 - p_2||^2),
\end{equation}
\vspace{-0.1em}
where $p_1$, $p_2$ are two points and $\lambda$ is a learnable parameter. With this kernel, an interpolated point is more greatly affected by closer points. The learnable RBF kernel is adopted by default in our approach.

\noindent\textbf{Plain Convolution} Plain convolution can also be used as a learnable interpolation kernel. However, negative convolution weights may result in a zero denominator for Eq.~(\ref{equ:windowed_interpolation_detail}). Therefore, we use the absolute value of the convolution weights in the denominator instead.

\noindent\textbf{Average Pooling} Average pooling is the simplest interpolation kernel. It assigns the same weight to all pixels in the interpolation window. We use this method as a baseline.

\vspace{-1em}
\subsection{Grid Prior}
\vspace{-0.5em}
With our stochastic sampling, the windowed sample set $\mathcal{R}_s^r$ at some unsampled points may be empty. To handle this situation, one way is to fill the features of these points as zeros. To allow interpolation at these points, we instead build an equal-interval sampling mask $\mathcal{M}_{grid}$ of stride $s$ and combine it with the mask $\mathcal{M}_{sample}$ generated by the sampling module:
$\mathcal{M} = \max(\mathcal{M}_{sample}, \mathcal{M}_{grid})$. The combined mask is used in the interpolation module, and we find experimentally that it does not affect the performance but better stabilizes the training process in comparison to zero filling, as shown in Sec.~\ref{sec:ablation}.

\vspace{-1em}
\subsection{Integration with Residual Block}
\label{sec:intergration_with_residual_block}
\input{figures/resblock.tex}
Our sampling and interpolation modules can be easily integrated with existing network architectures. Here, we use the residual block~\cite{he2016deep} as an example to show how these modules can be used. A natural way to insert them is by applying a separate sampling mask for each convolution (Fig.~\ref{fig:resblock} (c)). However, since conv3 is a pointwise convolution (kernel size is $1\times1$), its mask can be shared with conv2 without changing the output (Fig.~\ref{fig:resblock} (b)), meanwhile avoiding the computation for generating an extra mask. Another approach is to share a single mask among all convolutions within the residual block (Fig.~\ref{fig:resblock} (a)). However, since conv2 is not a pointwise convolution (kernel size is $3\times3$), sampling a point in conv2 
requires the corresponding $3\times3$ points from conv1 to be sampled, which in turn places strong conditions on the sampling of conv2 if conv1 and conv2 were to share the same mask. Thus, sharing one mask for all convolutions within a block is not an effective solution.
We found the two sampling mask approach (Fig.~\ref{fig:resblock} (b)) to be slightly better than others in experiments, so it is used by default.

%% file: figures/Pipeline-and-Sample-Comparison.tex
\begin{figure}[tbp]
  \centering
  \includegraphics[width=0.95\textwidth]{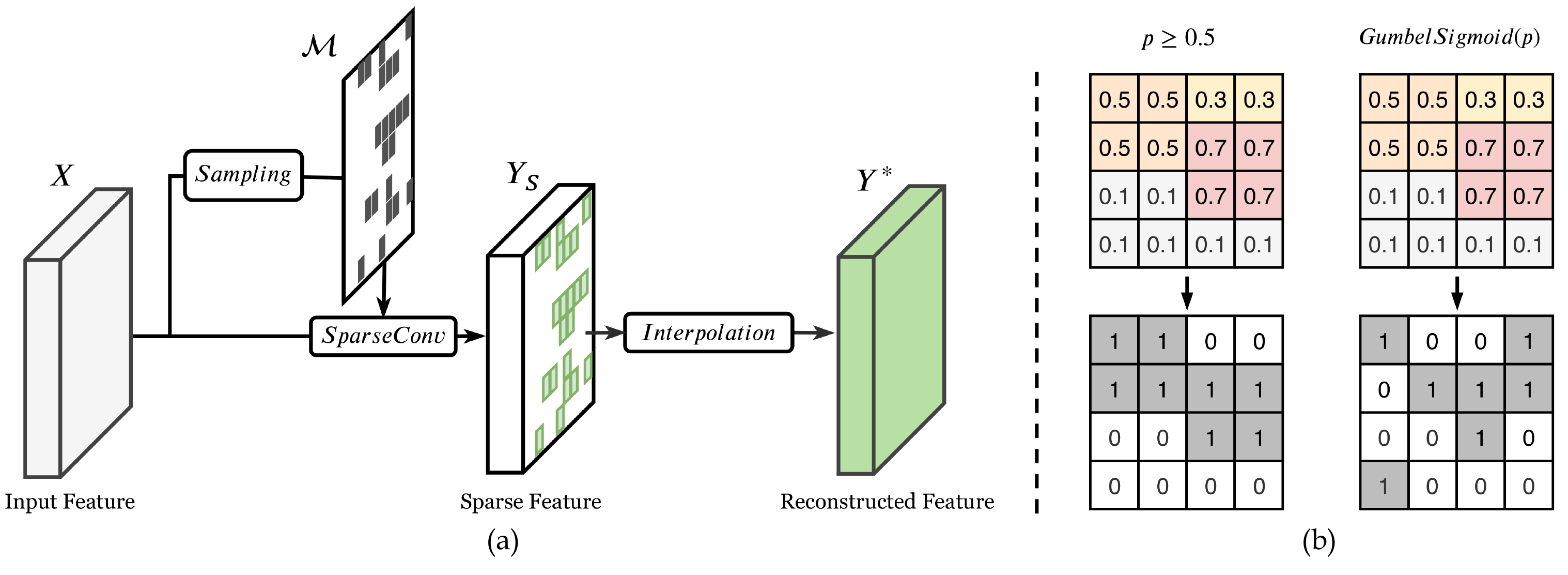}
  \vspace{-1.5em}
  \caption{(a) Stochastic sampling-interpolation network. The stochastic sampling module generates a sparse sampling mask $\mathcal{M}$ based on the input feature map $X$, and then calculates features only at the sampling points, forming a sparse feature map $Y_s$. The features of unsampled points are interpolated by the interpolation module to form the output feature map $Y^*$. (b)(Left) In deterministic sampling, points with the same confidence are either sampled altogether or not sampled at all. (Right) In stochastic sampling, a random subset of the points with the same confidence will be sampled, with a sampling density determined by their confidence.}
  \label{fig:pipline-and-sample-comparison}
  \vspace{-1em}
\end{figure}

%% file: figures/resblock.tex
\begin{figure}[tbp]
  \centering
  \includegraphics[width=0.75\textwidth]{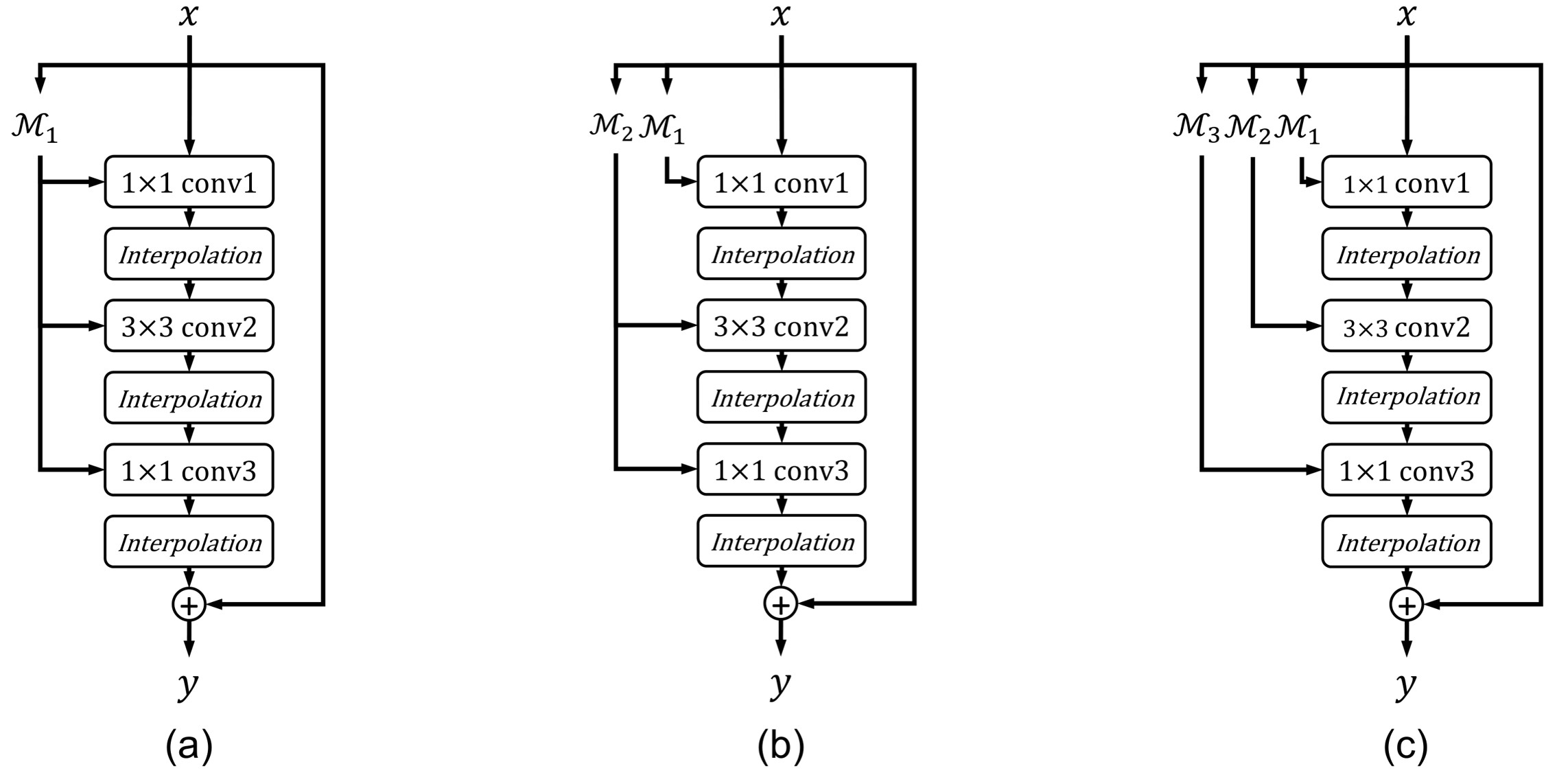}
  \vspace{-1.5em}
  \caption{Integration of our sampling and interpolation modules into a standard residual block. (a) Applying a single mask to all convolutions; (b) Applying a separate mask on conv1, and sharing a mask for conv2 and conv3. This approach is adopted as the default setting in our experiments; (c) Applying three separate masks for each of the convolutions.}
  \label{fig:resblock}
  \vspace{-1em}
\end{figure}

%% file: tex/experiments.tex
In this section, we validate our approach on three tasks: object detection, semantic segmentation, and image classification. Comparisons between our approach and other baseline models are conducted in terms of speed-accuracy tradeoff.
Average floating point operations (FLOPs) in the backbone network over the whole validation set is used to evaluate inference speed.
\vspace{-1em}
\subsection{Experimental settings}
\label{sec:exp_setting}
\vspace{-0.2em}
\noindent\textbf{Object Detection}
Our models are trained on the 118k images of the COCO 2017~\cite{lin2014coco} train set, and evaluated on the 5k images of the COCO 2017 validation set. The standard mean average precision (mAP) is used to measure accuracy. The baseline model is based on Faster R-CNN with Feature Pyramid Network (FPN)~\cite{lin2016feature} and deformable convolution~\cite{dai2017deformable,zhu2019deformable}. The other components and hyper-parameters are based on mmdetection~\cite{mmdetection2018}. More details are given in Appendix.

For our modules, the sparse sampling module and interpolation module are integrated into all the residual blocks as shown in Fig.~\ref{fig:resblock}~(b). The window size $r$ for interpolation is set to $7$, and the stride $s$ of the grid prior is set to $11$. During training, the parameter $\lambda$ used in the learnable RBF kernel is initialized to $3$, and the decay factor $\alpha$ of the stochastic sampling module is set to make the Gumbel-Softmax temperature parameter $\tau = 0.01$ at the end of training. During inference, we use the same $\tau$ to produce masks, and points with mask values below than $0.5$ are marked as unsampled.

\noindent\textbf{Semantic Segmentation}
Our models are trained on the $2975$ finely annotated images of the Cityscapes~\cite{cordts2016cityscapes} train set and evaluated on the $500$ images of the validation set. Accuracy is measured by the standard mean IoU. The baseline model is ResNet-50 based dilated FCN~\cite{long2015fully} with deformable convolution layers~\cite{dai2017deformable}. The implementation and hyper-parameters are based on the open-source implementation of TorchCV~\cite{you2019torchcv}. More details are given in Appendix. For our modules, the same hyper-parameters are used as in object detection.

\noindent\textbf{Image Classification}
Our models are trained on the ImageNet-1K training set, and follow both the training and inference settings of~\cite{dong2017more}. We choose ResNet-34 as our baseline for fair comparison. Following~\cite{dong2017more}, our sampling and interpolation modules are incorporated in all residual blocks, except for the first block in each stage. Since ResNet-34 is composed of basic blocks (two $3\times3$ convolutions), we apply separate masks for each convolution.

Different from the previous two tasks, experiments on ImageNet use a much lower image resolution (i.e. 224$\times$224). Thus, for our modules, we reduce the window size $r$ of the interpolation module to $5$, and the stride $s$ of the grid prior to $2$, while keeping other hyper-parameters the same as in object detection.

\vspace{-0.7em}
\subsection{Ablation Study}
\vspace{-0.2em}
\label{sec:ablation}
We validate several design choices on the COCO2017 object detection benchmark. All the models are trained and evaluated on images with a shorter side of 1000 pixels and with the sparse loss weight set to 0.1. 

\input{table/ablation.tex}

\noindent\textbf{Different interpolation kernels} We first compare the different interpolation kernels mentioned in Sec.~\ref{sec:interpolation_module}: \textit{learnable RBF kernel}, \textit{plain convolution} and \textit{average pooling}. Results are shown in Table~\ref{table.ablation_interpolation_kernels}, with the sparse loss weight in the training phase set so that the three kernels yield similar accuracy. It can be seen that the \textit{learnable RBF kernel} consumes fewer FLops.

\noindent\textbf{Effect of removing interpolation} We further study the effect of removing the interpolation module by replacing it with \textit{reusing features} of the previous layer~\cite{figurnov2017spatially} (see Eq.~(\ref{equ:reuse_feature})) and directly \textit{filling zeros}~\cite{dong2017more} (see Eq.~(\ref{equ:fill_zero})). Results are shown in Table~\ref{table.ablation_effect_interpolation}. \textit{Reusing features} achieves 42.1 mAP with 226.1 GFLOPs, while \textit{Filling zeros} obtains 42.0 mAP with 164.9 GFLOPs. Both of these methods consume noticeably more computation than our interpolation module with similar or worse accuracy, which indicates that they are inferior to our method in terms of FLOPs-accuracy tradeoff.

\noindent\textbf{Effect of Grid Prior} We compare performance under different grid stride $s$ or by removing the grid prior. Results are shown in Table~\ref{table.ablation_grid}. We found the performance in different settings to be similar, but the training process is sometimes not stable at $s=13$ and without the grid prior. Therefore, we choose $s=11$ as our default setting.

\input{figures/Det-Seg-Result}
\vspace{-1em}
\subsection{Object Detection}
\label{sec:object_detection}
\vspace{-0.5em}
We compared our method to other baselines on the COCO2017 object detection benchmark. To better illustrate the FLOPs-accuracy tradeoff under different parameters and settings, we display charts rather than tables and present the numerical results in Appendix.

\noindent\textbf{Uniform sampling} The direct way to sample uniformly is by using a mask with sampling points at equal intervals. However, since the interval must be integer, the minimal interval would be 2, which limits its feasibility for handling arbitrary sampling ratios. Thus, instead of using an equal-interval sampling mask, we choose to downsample the input images and not include our modules, which can also be seen as a uniform sampling method. We compared these two approaches with a downsampling rate of 2, and experimental results show the performance of these two approaches to be very similar\footnote{38.5 mAP for downsampled images vs.~38.7 mAP for equal-interval sampling masks.}.

For uniform sampling, we resize the shorter sides of input images to \{1000, 800, 600, 500\} to draw the FLOPs-accuracy curve. For our method, the curve is drawn according to different sparse loss weights, i.e. \{0.2, 0.1, 0.05, 0.02\}. Results are shown in Fig.~\ref{fig:det_seg_results}(a). For our baseline without any sampling or interpolation modules, which is trained and evaluated on images with a shorter side of 1000 pixels, it obtains 43.4 mAP with 289.5 GFLOPs. In comparison, our stochastic sampling method performs at 43.3 mAP with only 160.4 GFLOPs, saving nearly half of the computation cost with no accuracy drop.

\noindent\textbf{Deterministic sampling}
We next compare our stochastic sampling strategy to deterministic sampling methods. For fair comparison, we only replace the sampling module, while leaving the other parts unchanged. Two deterministic sampling methods are examined:
\begin{itemize}

\item ReLU Gating: Similar to LCCL~\cite{dong2017more}, a ReLU function is applied on sampling confidence to generate a sparse mask, by trimming values smaller than 0 during training and inference. The sampling confidence is generated by a $3\times3$ convolution with a 1-dimensional output. Since the output of ReLU is not binary, Eq.~(\ref{equ:soft-output}) is used to calculate the full output feature.

\item Deterministic Gumbel-Softmax: Eliminating the noise $g$ in Gumbel-Softmax naturally results in deterministic sampling, as described in Sec.~\ref{sec:stochastic_sampling_module}, and is approximately equal to the Softmax function with a temperature term.
\end{itemize}

\noindent For our stochastic sampling and the two deterministic sampling methods, the FLOPs-accuracy curve is drawn according to different sparse loss weights, i.e. \{0.2, 0.1, 0.05, 0.02\}. Compared to \textit{ReLU Gating}, our stochastic sampling achieves clearly better performance than the ReLU based approach, especially at lower levels of computation. \textit{Deterministic Gumbel-Softmax} performs better than \textit{ReLU Gating}, but is still worse than stochastic sampling by a large margin.

\noindent\textbf{Smaller backbones} Replacing a large backbone with a smaller backbone is a common method for reducing computation. Thus, we also compare our method (based on ResNet-101) with a baseline using ResNet-50 as the smaller backbone. ResNet-50 achieves 41.2 mAP with 149.2 GFLOPs, which indicates accuracy much worse than that of our method with similar computation costs.

\vspace{-1em}
\subsection{Semantic Segmentation}
We also conduct experiments on the Cityscapes benchmark for semantic segmentation. Unless otherwise specified, all the models are trained and evaluated on images with a shorter side of 1024 pixels. We first compare our content-aware stochastic sampling to \textit{uniform sampling} and deterministic sampling. For deterministic sampling, we choose \textit{deterministic Gumbel-Softmax} for comparison because of its better performance exhibited on the object detection task. 

Results are shown in Fig.~\ref{fig:det_seg_results}(b) and present the numerical results in Appendix. For our method and \textit{deterministic Gumbel-Softmax}, we draw curves according to different sparse loss weights \{0.3, 0.2, 0.1, 0.05\}. For \textit{uniform sampling}, we resize the shorter sides of input images to \{1024, 896, 736, 512\}. The original baseline with a shorter side of 1024 pixels obtains 80.8 mean IoU with 920.6 GFLOPs. In comparison, our method obtains 80.6 mean IoU with only 373.2 GFLOPs, saving nearly 60\% of the computation cost with almost no accuracy drop. Compared with other sampling methods, i.e. \textit{uniform sampling} and \textit{deterministic Gumbel-Softmax}, our method clearly achieves a better FLOPs-accuracy tradeoff. 

\vspace{-1em}
\input{table/cls_res.tex}
\subsection{Image Classification}
\label{sec:image_cls}
We also compare our method to other state-of-the-art methods~\cite{dong2017more,li2017pruning,he2018soft,he2019filter} for reducing computation on the ImageNet-1K image classification benchmark. Similar to our method, LCCL~\cite{dong2017more} explores sparsity in the spatial domain, while PFEC~\cite{li2017pruning}, SFP~\cite{he2018soft} and FPGM~\cite{he2019filter} reduce computation by model pruning. Results are presented in Table~\ref{table.cls_res}. Our models are trained under different sparse loss weights, 0.01 and 0.015, to reach accuracy or FLOPs similar to other methods for fair comparison. Compared with LCCL~\cite{dong2017more} and PFEC~\cite{li2017pruning}, our approach achieves comparable accuracy with less FLOPs. Compared with SFP~\cite{he2018soft} and FPGM~\cite{he2019filter}, our approach obtains a smaller accuracy drop with similar FLOPs. 

We further remove the interpolation module from our method and fill the features of unsampled points with 0. Results show that removing interpolation does not affect performance on the ImageNet validation set. This is inconsistent with object detection and semantic segmentation. We believe that this is because the classification network is focused on extracting global feature representations. Therefore, as long as the features of certain key points are calculated and preserved, the global features will not be affected and the performance will not be hurt. In other words, in the image classification task, it is not important to reconstruct the features of unsampled points by interpolation.

\input{figures/gamma_sparsity.tex}
\vspace{-1em}
\subsection{Analysis of sampling and interpolation modules}
\label{sec:analysis}
In this section, we further study the relationship between the sampling and interpolation modules. Specifically, we analyze the relationship between sampling sparsity and the parameter $\lambda$ in the RBF interpolation module. A larger $\lambda$ indicates a sharper interpolation kernel, and when $\lambda > 3$, the RBF interpolation kernel approximates an identity kernel, for which there is no interpolation.

Results are shown in Fig.~\ref{fig:gamma_sparsity}. For object detection (see Fig.~\ref{fig:gamma_sparsity}~(a)) and semantic segmentation (see Fig.~\ref{fig:gamma_sparsity}~(b)), sparsity and $\lambda$ show a strong negative correlation. When sparsity is high, the $\lambda$ is usually small, which means sampled points far away from the unsampled points also contribute greatly to the interpolation. However, for image classification (see Fig.~\ref{fig:gamma_sparsity}~(c)), this correlation has not been observed.  $\lambda$ is quite large in most cases, which means the effect of the interpolation module is limited. This phenomenon is consistent with the experimental results of the ``w/o Interp'' entry in Table~\ref{table.cls_res}, that the results without interpolation are almost identical to our full model, further indicating that interpolation is not important for image classification. 

Another observation is that the $\lambda$ of conv1 are consistently smaller than $\lambda$ of conv2 and conv3 in object detection and semantic segmentation. The reason is still unclear but we suspect that this phenomenon may be related to the receptive field of operators. 

\input{table/cpu_runtime.tex}

\vspace{-1em}
\subsection{Realistic run-time on CPU}
Our method achieves good theoretical speed-accuracy trade-offs. In this section, we present a preliminary evaluation of the realistic run-time of the backbone network on the CPU. According to Eq.~(2), the mask $\mathcal{M}$ is calculated before computing $Y$ to decide which points in $Y$ need to calculated. In order to show the realistic speedup of our method under different computing resources, we conducted experiments in two different hardware environments: a workstation (E5-2650 v2, 256G RAM and Ubuntu 16.04 OS) and a laptop (I7-6650U, 16G RAM and Ubuntu 16.04 OS). For fair comparison, we replace all convolutions by our implementation for all models, and enable multi-threading by default (32 threads for E5-2650 v2 and 4 threads for I7-6650U). Results on object detection are shown in Table~\ref{table.cpu_runtime}. There is a gap between the theoretical speedup and the realistic speedup, but the gap for laptop (I7-6650U) is smaller than for workstation (E5-2650 v2). The main reason is that the workstation has better computing speed and more cores (E5-2650 v2 has 8 cores and I7-6650U has 2 cores), but has a memory access bottleneck. This suggests that our method is more suitable for devices with less computation speed but relatively faster IO speed, such as mobile or edge computing devices. 

In addition, some works~\cite{ren2018sbnet,parashar2017scnn} have developed general techniques to accelerate sparse convolution based methods. For example, SCNN~\cite{parashar2017scnn} designed a hardware accelerator for sparse convolution and demonstrate that ideal speedup of sparse convolution is achievable in such devices. SBNet~\cite{ren2018sbnet} is a general method to accelerate sparse convolution in GPU. These general techniques of accelerating sparse convolution are compatible with our method, and can further close the gap between theoretical and actual speedup in real applications.

\section{Compatibility with Pruning Method}
\vspace{-0.5em}
High computational costs are alleviated in our work by exploring spatial redundancy, an approach that differs from other techniques such as model pruning. For further verification that our method is compatible with pruned models, we applied the global unstructured pruning method provided in the official PyTorch implementation (\texttt{torch.nn.utils.prune.global\_unstructured}) on the backbone network, but exclude the offset convolution layer used in deformable convolution and the mask prediction layer of our method. A comparison between the baseline model and our model on the COCO object detection benchmark is shown in Table~\ref{table.compatibility}. Since our method introduces additional parameters to predict the sampling positions, the number of parameters is slightly larger than that of the baseline model. The results show that at various pruning levels, the mAP performance of our model is comparable to that of the baseline model. Moreover, it is seen that our method consistently uses less FLOPs than the baseline model under similar mAP. These results indicate that our method is complementary to and compatible with the pruning method.
\input{table/compatibility_prune}

\vspace{-0.7em}
\section{Conclusion}
\vspace{-0.7em}
A method for reducing computation in convolutional networks was proposed that exploits the intrinsic sparsity and spatial redundancy in feature maps. We present a stochastic sampling and interpolation scheme to avoid expensive computation at spatial locations that can be effectively interpolated. To overcome the challenge of training binary decision variables for representing discrete sampling locations, Gumbel-Softmax is introduced to our sampling module. The effectiveness of this approach is verified on a variety of computer vision tasks.

%% file: table/ablation.tex
\vspace{-1em}
\begin{table}[H]
\small
\caption{Comparison of different interpolation kernels on COCO2017 validation}
\vspace{-1.5em}
\begin{center}
\resizebox{0.48\linewidth}{!}{
\begin{tabular}{c|c|c|c}
\hline
\tabincell{c}{} & \tabincell{c}{Avg Pool} & \tabincell{c}{Plain Conv} & \tabincell{c}{RBF Kernel}\\
\hline
\hline
mAP & 41.8 & 41.9 & 42.0 \\
\hline
GFlops & 110.0 & 109.8 & 96.6 \\
\hline
\end{tabular}
}
\end{center}
\label{table.ablation_interpolation_kernels}
\vspace{-5em}
\end{table}

\begin{table}[H]
\small
\caption{Validation of the interpolation module on COCO2017 validation}
\vspace{-1.5em}
\begin{center}
\resizebox{0.46\linewidth}{!}{
\begin{tabular}{c|c|c|c}
\hline
\tabincell{c}{} & \tabincell{c}{Fill Zeros} & \tabincell{c}{Reuse Features} & \tabincell{c}{Ours}\\
\hline
\hline
mAP & 42.0 & 42.1 & 42.0 \\
\hline
GFlops & 164.9 & 226.1 & 96.6 \\
\hline
\end{tabular}
}
\end{center}
\label{table.ablation_effect_interpolation}
\vspace{-5em}
\end{table}

\begin{table}[H]
\small
\caption{Comparison of different grid prior settings on COCO2017 validation}
\vspace{-1.5em}
\begin{center}
\resizebox{0.5\linewidth}{!}{
\begin{tabular}{c|c|c|c|c}
\hline
\tabincell{c}{} & \tabincell{c}{$s=9$} & \tabincell{c}{$s=11$} & \tabincell{c}{$s=13$} & \tabincell{c}{w/o Grid Prior} \\
\hline
\hline
mAP & 41.9 & 42.0 & 41.8 & 41.8\\
GFlops & 95.4 & 96.6 & 95.3 & 95.0 \\
\hline
\end{tabular}
}
\end{center}
\vspace{-2em}
\label{table.ablation_grid}
\end{table}

%% file: figures/Det-Seg-Result.tex
\begin{figure}[tbp]
  \centering
  \includegraphics[width=1.0\textwidth]{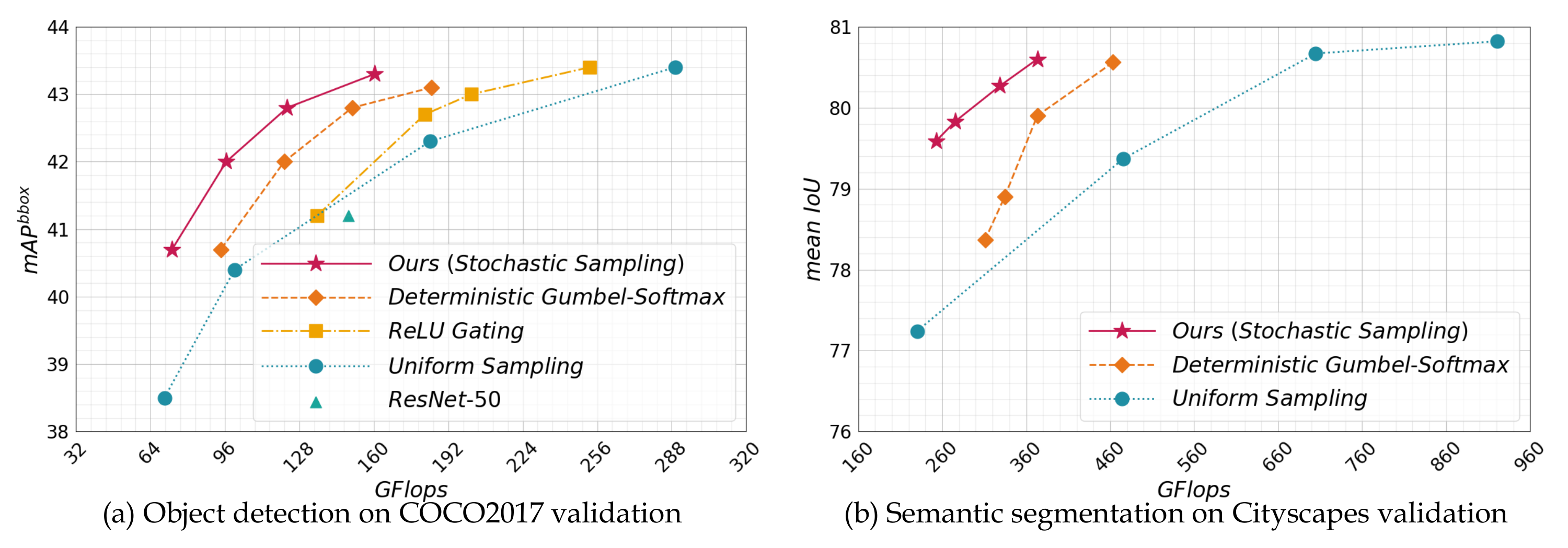}
  \vspace{-1.2em}
  \caption{(a) Tradeoff curves for different sampling methods on object detection (COCO2017 validation). Curve of ``Uniform Sampling'' is drawn from various resolutions, and others are drawn from various sparse loss weights (with shorter side of 1000 pixels). (b) Experiments on the Cityscapes semantic segmentation benchmark. Except for ``Uniform Sampling'', all other models are trained and evaluated on images with a shorter side of 1024 pixels.}
  \label{fig:det_seg_results}
  \vspace{-1.5em}
\end{figure}

%% file: table/cls_res.tex
\setlength{\tabcolsep}{2pt}
\renewcommand{\arraystretch}{1.0}
\begin{table}[t]
\small
\caption{Performance comparison on the ImageNet validation set. All the methods are based on ResNet-34. Our models are trained with a loss weight of 0.01 and 0.015 to achieve accuracy or FLOPs similar to other methods for fair comparison. ``w/o Interp'' indicates removing the interpolation module and filling the features of unsampled positions with 0}
\begin{center}
\resizebox{0.7\linewidth}{!}{
\begin{tabular}{c|c|c|c|c}
\hline
\tabincell{c}{Method} & \tabincell{c}{Type} & \tabincell{c}{\small{Top-1/Top-5} \\ \small{Acc Drop($\%$)}} & \tabincell{c}{FLOPs} & \tabincell{c}{Speedup} \\
\hline
\hline
LCCL~\cite{dong2017more} & spatial & $0.43/0.17$ & $2.7\times10^9$ & $24.8\%$ \\
\hline
PFEC~\cite{li2017pruning} & pruning & $1.06/$- & $2.7\times10^9$ & $24.2\%$ \\
\hline
SFP~\cite{he2018soft} & pruning & $2.09/1.29$ & $2.2\times10^9$ & $41.1\%$ \\
\hline
FPGM~\cite{he2019filter} & pruning & $1.29/0.54$ & $2.2\times10^9$ & $41.1\%$ \\
\hline
\hline
Ours(0.01) & spatial & $0.45/0.20$ & $2.53\times10^9$ & $30.8\%$\\
\hline
Ours(0.015) & spatial & $1.19/0.47$ & $2.16\times10^9$ & $42.4\%$\\
\hline
w/o Interp(0.015) & spatial & $1.07/0.46$ & $2.21\times10^9$ & $41.0\%$ \\
\hline
\end{tabular}}
\end{center}
\label{table.cls_res}
\end{table}

%% file: figures/gamma_sparsity.tex
\begin{figure}[tbp]
\centering 
\hspace{-1.6em}
\subfigure[\scriptsize Object Detection]{
\includegraphics[width=0.27\linewidth]{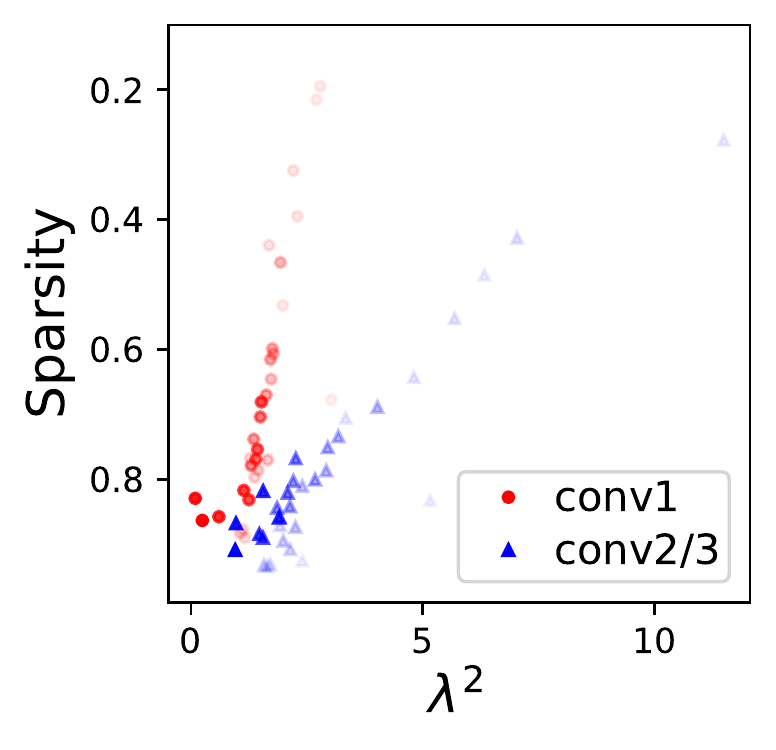}
}
\hspace{-1.23em}
\subfigure[\scriptsize Semantic Segmentation]{
\includegraphics[width=0.27\linewidth]{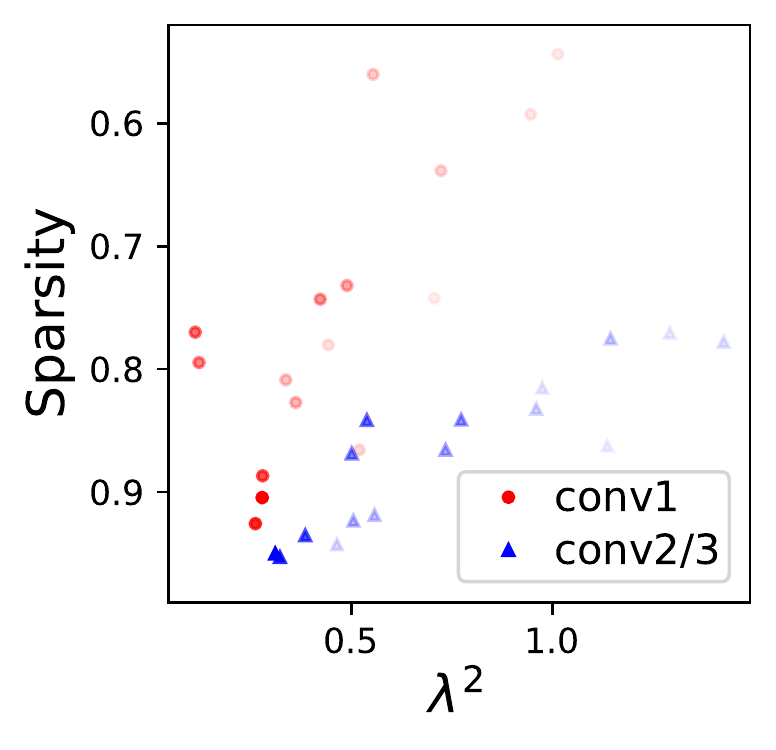} 
}
\hspace{-1.22em}
\subfigure[\scriptsize Image Classification]{
\includegraphics[width=0.27\linewidth]{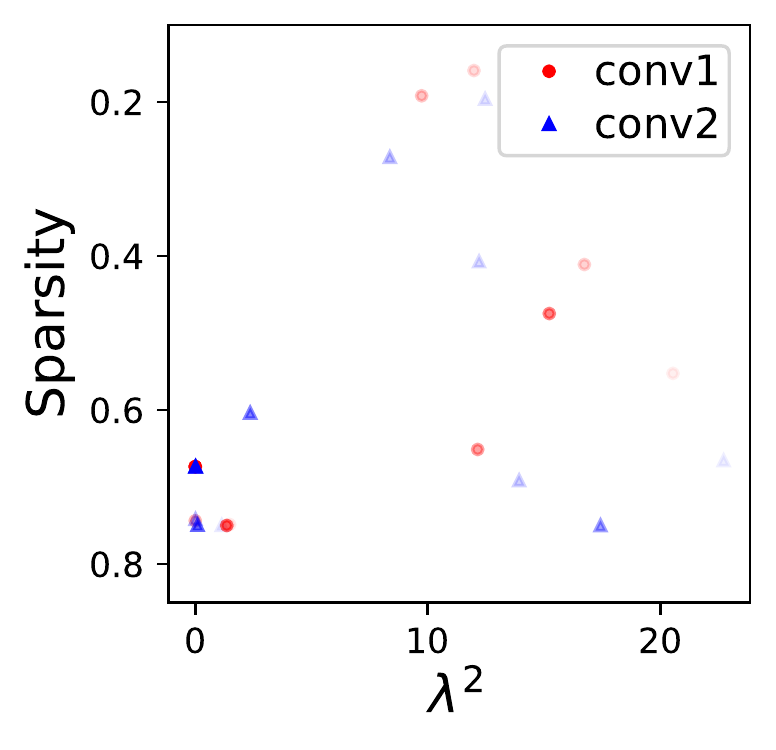} 
}
\vspace{-1.0em}
\caption{Relationship between interpolation parameter $\lambda$ and sparsity. The transparency of each point represents its location in the network, where a darker color indicates greater depth. A larger $\lambda$ yields a sharper interpolation kernel; when $\lambda > 3$, it approximates an identity kernel.} 
\vspace{-1em}
\label{fig:gamma_sparsity} 
\end{figure} 

%% file: table/cpu_runtime.tex
\setlength{\tabcolsep}{2pt}
\renewcommand{\arraystretch}{1.0}
\begin{table}[t]
\small
\caption{Comparison of theoretical and realistic speedups on E5-2650 and I7-6650U. Baseline model is trained and evaluated on images with a shorter side of 1000 pixels. The CPU run-time is calculated on the COCO2017 validation set}
\begin{center}
\resizebox{0.95\linewidth}{!}{
\begin{tabular}{c|c|c|c|c|c|c|c}
\hline
\multirow{2}{*}{Model} & \multirow{2}{*}{mAP} & \multirow{2}{*}{GFLOPs} &  
\multicolumn{2}{c|}{Runtime(s/img)} & \multicolumn{2}{c|}{Real Speedup} &  \multirow{2}{*}{Theo Speedup} \\
\cline{4-7}
& & &  E5-2650 v2 & I7-6650U & E5-2650 v2 & I7-6650U & \\
\hline
\hline
Baseline & 43.4 & 289.5 & 4.9 & 10.5 & - & - & - \\
\hline 
Our(0.02) & 43.3 & 160.4 & 4.0 & 7.9 & 1.23 & 1.33 & 1.80 \\
\hline
Our(0.05) & 42.8 & 122.8 & 3.5 & 6.3 & 1.40 & 1.67 & 2.36 \\
\hline
Our(0.1) & 42.0 & 96.6 & 3.2 & 5.3 & 1.53 & 1.98 & 3.00 \\
\hline
Our(0.2) & 40.7 & 73.3 & 2.7 & 4.4 & 1.81 & 2.39 & 3.95 \\
\hline
\end{tabular}
}
\end{center}
\vspace{-2em}
\label{table.cpu_runtime}
\end{table}

%% file: table/compatibility_prune.tex
\setlength{\tabcolsep}{2pt}
\renewcommand{\arraystretch}{1.0}
\begin{table}[t]
\small
\caption{Studying the compatibility with pruning method. We applied the global unstructured pruning method on baseline model and our model (sparse loss weight of 0.02) with various pruning ratios.}
\begin{center}
\resizebox{0.8\linewidth}{!}{
\begin{tabular}{c|c|c|c|c|c|c}
\hline
\multirow{2}{*}{Prune Ratio} &  
\multicolumn{3}{c|}{Baseline} & \multicolumn{3}{c}{Ours(0.02)}\\
\cline{2-7}
& params(M) & GFLOPs & mAP & params(M) & GFLOPs & mAP \\
\hline
\hline
0.00 & 47.25 & 289.5 & 43.4 & 47.80 & 160.4 & 43.3 \\ 
\hline
0.50 & 26.05 & 166.4& 42.8 & 26.60 & 96.8 & 42.7 \\ 
\hline
0.55 & 23.93 & 155.5 & 42.1 & 24.48 & 91.0 & 42.2 \\ 
\hline
0.60 & 21.81 & 144.5 & 40.9 & 22.36 & 85.0 & 41.1 \\ 
\hline
0.66 & 19.69 & 133.3 & 38.4 & 20.24 & 78.8 & 38.8 \\ 
\hline
0.79 & 17.57 & 122.0 & 32.7 & 18.12 & 72.5 & 34.0 \\ 
\hline
0.75 & 15.45 & 110.3 & 20.3 & 16.0 & 65.7 & 23.4 \\ 
\hline
\end{tabular}
}
\end{center}
\vspace{-2.5em}
\label{table.compatibility}
\end{table}

%% file: table/object_res.tex
\setlength{\tabcolsep}{2pt}
\renewcommand{\arraystretch}{1.0}

\begin{table}[t]
\small
\caption{The numerical results of Fig.~\ref{fig:det_seg_results} (a) in the main paper. Experiments are conducted on object detection (COCO2017 validation)}
\begin{center}
\resizebox{0.55\linewidth}{!}{
\begin{tabular}{c|c|c}
\hline
\tabincell{c}{Model} & \tabincell{c}{mAP} & \tabincell{c}{GFLOPs} \\
\hline
\hline
ResNet-50(1000 px) & 41.2 & 149.2 \\
\hline
\hline
Uniform Sampling(1000 px) & 43.4 & 289.5 \\
\hline
Uniform Sampling(800 px) & 42.3 & 184.1 \\
\hline
Uniform Sampling(600 px) & 40.4 & 100.2 \\
\hline
Uniform Sampling(500 px) & 38.5 & 70.0 \\
\hline
\hline
Determ Gumble-Softmax(0.02) & 43.1 & 184.9 \\ 
\hline
Determ Gumble-Softmax(0.05) & 42.8 & 150.7 \\ 
\hline
Determ Gumble-Softmax(0.1) & 42.0 & 121.5 \\ 
\hline
Determ Gumble-Softmax(0.2) & 40.7 & 94.3 \\ 
\hline
\hline
ReLU(0.02) & 43.4 & 252.8 \\ 
\hline
ReLU(0.05) & 43.0 & 202.0 \\ 
\hline
ReLU(0.1) & 42.7 & 181.9 \\ 
\hline
ReLU(0.2) & 41.2 & 135.6 \\ 
\hline
\hline
Ours(0.02) & 43.3 & 160.4 \\
\hline
Ours(0.05) & 42.8 & 122.8 \\
\hline
Ours(0.1) & 42.0 & 96.6 \\
\hline
Ours(0.2) & 40.7 & 73.3 \\
\hline
\end{tabular}}
\end{center}
\vspace{-2em}
\label{table.object_res}
\end{table}

%% file: table/seg_res.tex
\setlength{\tabcolsep}{2pt}
\renewcommand{\arraystretch}{1.0}
\begin{table}[t]
\small
\caption{Numerical results of Fig.~\ref{fig:det_seg_results} (b) in the main paper. Experiments are conducted on semantic segmentation (Cityscapes validation)}
\begin{center}
\resizebox{0.57\linewidth}{!}{
\begin{tabular}{c|c|c}
\hline
\tabincell{c}{Model} & \tabincell{c}{mean IoU} & \tabincell{c}{GFLOPs} \\
\hline
\hline
Uniform Sampling(1024 px) & 80.82 & 920.6 \\
\hline
Uniform Sampling(896 px) & 80.67 & 704.8 \\
\hline
Uniform Sampling(736 px) & 79.37 & 475.6 \\
\hline
Uniform Sampling(512 px) & 77.24 & 230.1 \\
\hline
\hline
Determ Gumble-Softmax(0.05) & 80.56 & 463.4 \\ 
\hline
Determ Gumble-Softmax(0.1) & 79.90 & 373.2 \\ 
\hline
Determ Gumble-Softmax(0.2) & 78.90 & 334.4 \\ 
\hline
Determ Gumble-Softmax(0.3) & 78.37 & 311.1 \\ 
\hline
\hline
Ours(0.05) & 80.60 & 373.2 \\
\hline
Ours(0.1) & 80.27 & 328.8 \\
\hline
Ours(0.2) & 79.83 & 275.4 \\
\hline
Ours(0.3) & 79.59 & 252.9 \\
\hline
\end{tabular}}
\end{center}
\vspace{-1em}
\label{table.seg_res}
\end{table}

%% file: table/inference_stability.tex
\setlength{\tabcolsep}{2pt}
\renewcommand{\arraystretch}{1.0}
\begin{table}[t]
\small
\caption{Evaluation of inference stability on object detection (COCO2017 validation)}
\begin{center}
\resizebox{0.55\linewidth}{!}{
\begin{tabular}{c|c|c|c}
\hline
\tabincell{c}{Loss Weight} & \tabincell{c}{Grid Prior} & \tabincell{c}{mAP \\ mean / std} & \tabincell{c}{GFLOPs \\ mean / std} \\
\hline
\hline
0.02 & \checkmark & $43.28/0.02$ & $160.35/0.02$ \\
\hline
0.05 & \checkmark & $42.81/0.02$ & $122.79/0.03$ \\
\hline
0.1 & \checkmark & $41.98/0.02$ & $96.64/0.03$ \\
\hline
0.2 & \checkmark & $40.69/0.03$ & $73.25/0.02$ \\
\hline
\hline
0.02 &  & $42.61/0.11$ & $156.99/0.02$ \\
\hline
0.05 &  & $41.25/0.05$ & $113.47/0.02$ \\
\hline
0.1 &  & $38.59/0.09$ & $85.87/0.02$ \\
\hline
0.2 &  & $32.63/0.08$ & $65.03/0.03$ \\
\hline
\end{tabular}}
\end{center}
\vspace{-1.5em}
\label{table.inf_stab}
\end{table}

%% file: figures/resnext_det.tex
\begin{figure}[tbp]
  \centering
  \includegraphics[width=0.5\textwidth]{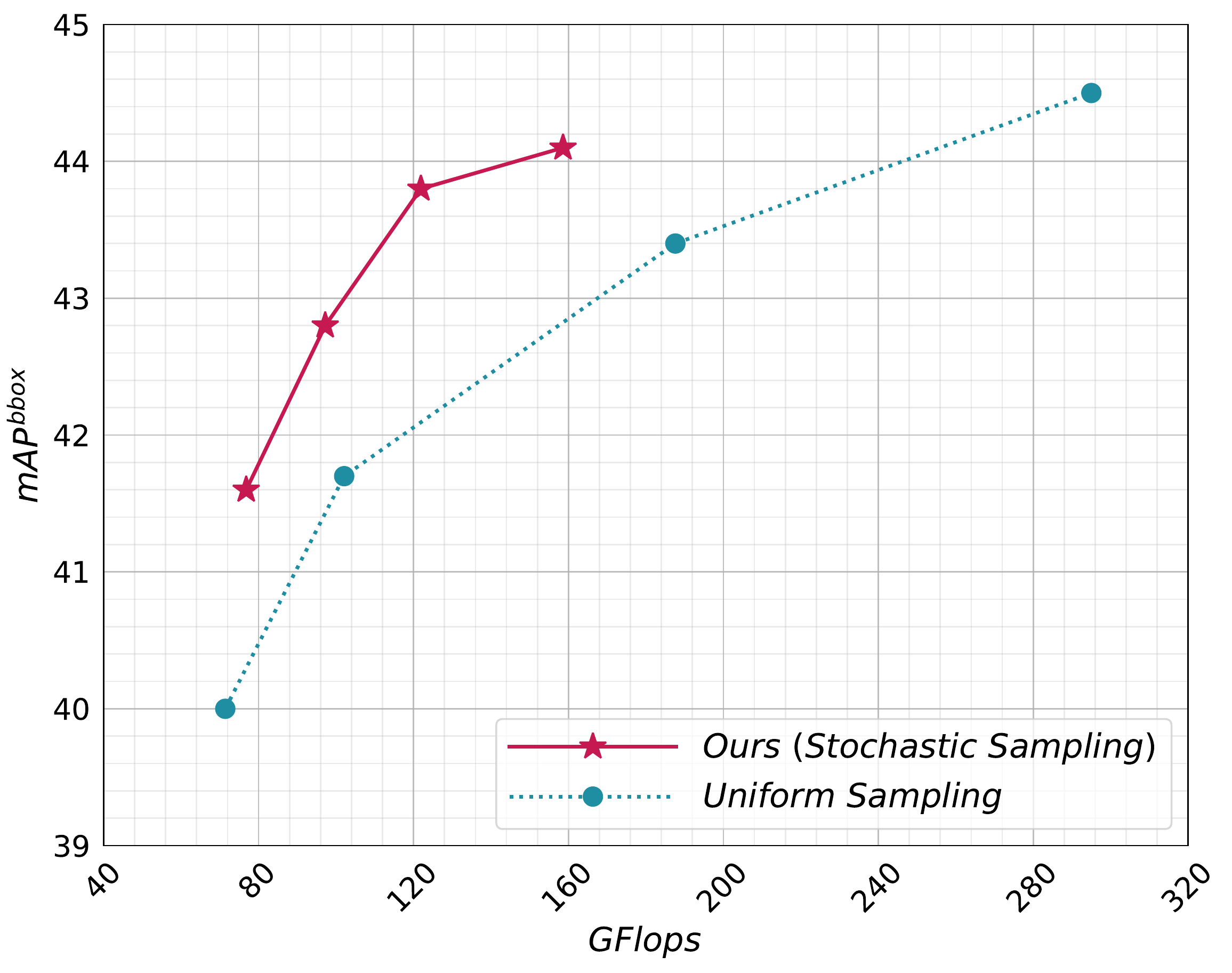}
  \vspace{-0.5em}
  \caption{Comparison of our method and uniform sampling on object detection (COCO2017 validation). ResNeXt is chosen as the backbone model. Curve of ``Uniform Sampling'' is drawn from various resolutions. Curve of our method is drawn from various sparse loss weights (with shorter side of 1000 pixels).}
  \label{fig:resnext_det}
\end{figure}

%% file: table/resnext_res.tex
\setlength{\tabcolsep}{2pt}
\renewcommand{\arraystretch}{1.0}
\begin{table}[t]
\small
\caption{Numerical results of Fig.~\ref{fig:resnext_det}. ResNeXt is chosen as the backbone model. Experiments are conducted on object detection (COCO2017 validation)}
\begin{center}
\resizebox{0.55\linewidth}{!}{
\begin{tabular}{c|c|c}
\hline
\tabincell{c}{Model} & \tabincell{c}{mAP} & \tabincell{c}{GFLOPs} \\
\hline
\hline
Uniform Sampling(1000 px) & 44.5 & 295.0 \\
\hline
Uniform Sampling(800 px) & 43.4 & 187.6 \\
\hline
Uniform Sampling(600 px) & 41.7 & 102.1 \\
\hline
Uniform Sampling(500 px) & 40.0 & 71.4 \\
\hline
\hline
Ours(0.02) & 44.1 & 158.6 \\
\hline
Ours(0.05) & 43.8 & 121.9 \\
\hline
Ours(0.1) & 42.8 & 97.2 \\
\hline
Ours(0.2) & 41.6 & 76.8 \\
\hline
\end{tabular}}
\end{center}
\vspace{-2em}
\label{table.resnext_res}
\end{table}

%% file: figures/sparsity_diff_layers.tex
\begin{figure*}[tbp]
  \centering
    \subfigure[Model trained on loss weight 0.02]{
    \includegraphics[width=0.95\linewidth]{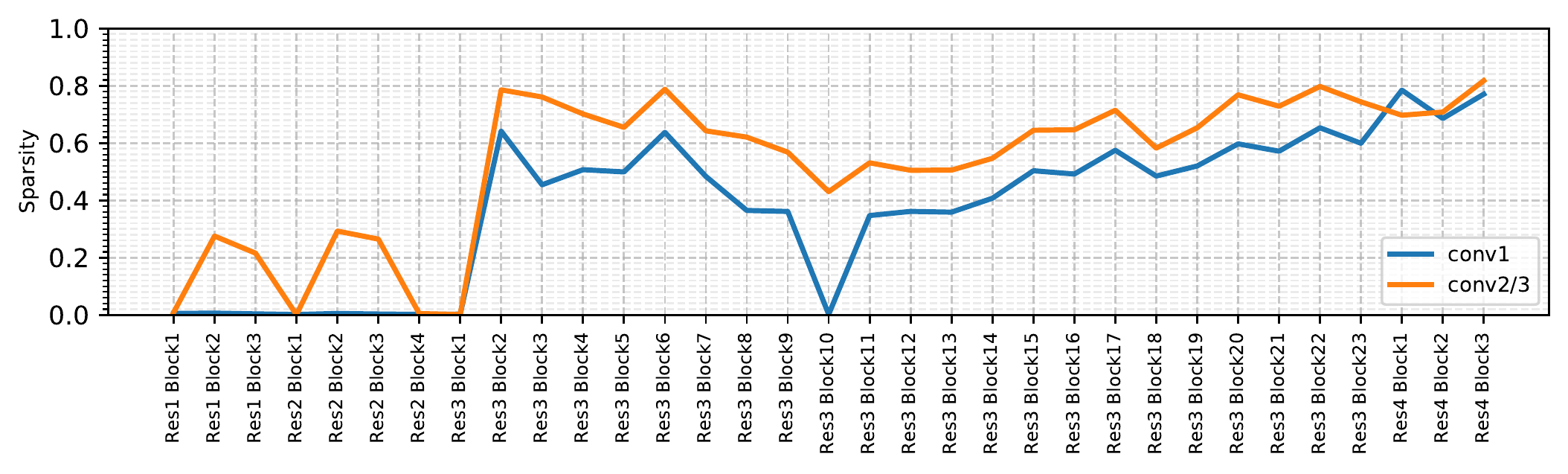}
    }
    \vspace{-0.5em}
    \subfigure[Model trained on loss weight 0.05]{
    \includegraphics[width=0.95\linewidth]{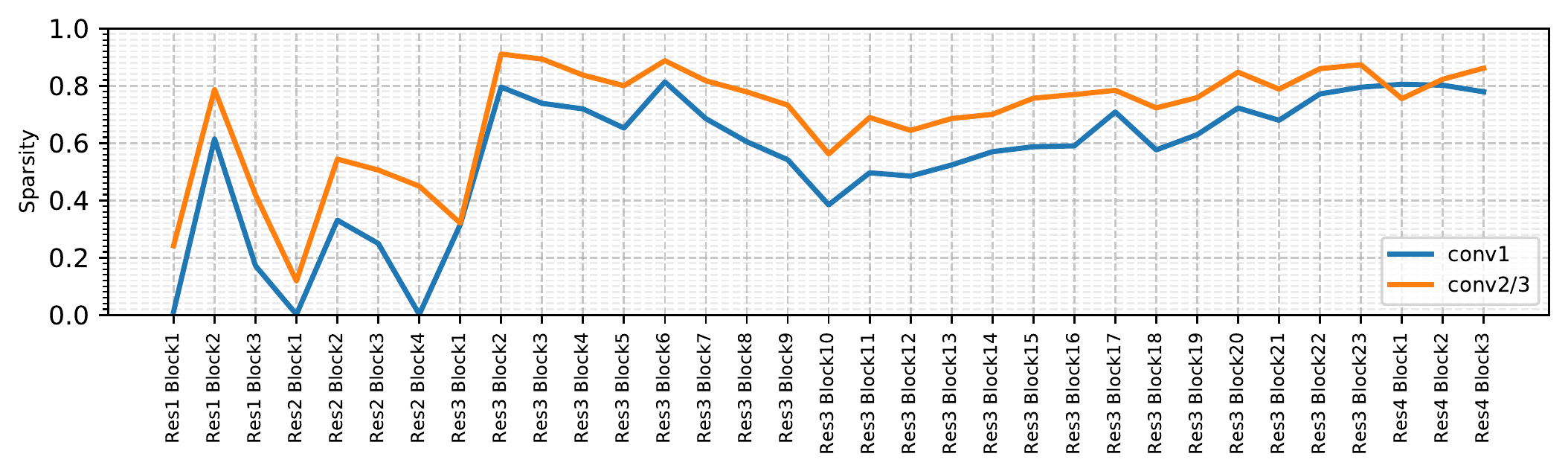} 
    }
    \vspace{-0.5em}
    \subfigure[Model trained on loss weight 0.10]{
    \includegraphics[width=0.95\linewidth]{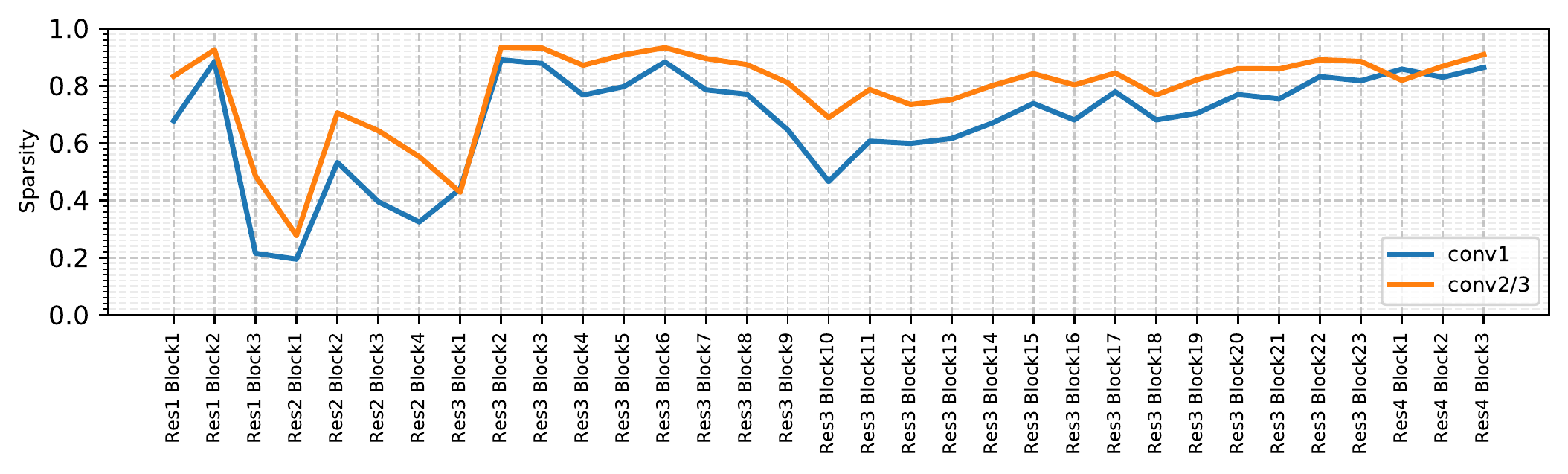} 
    }
    \vspace{-0.5em}
    \subfigure[Model trained on loss weight 0.20]{
    \includegraphics[width=0.95\linewidth]{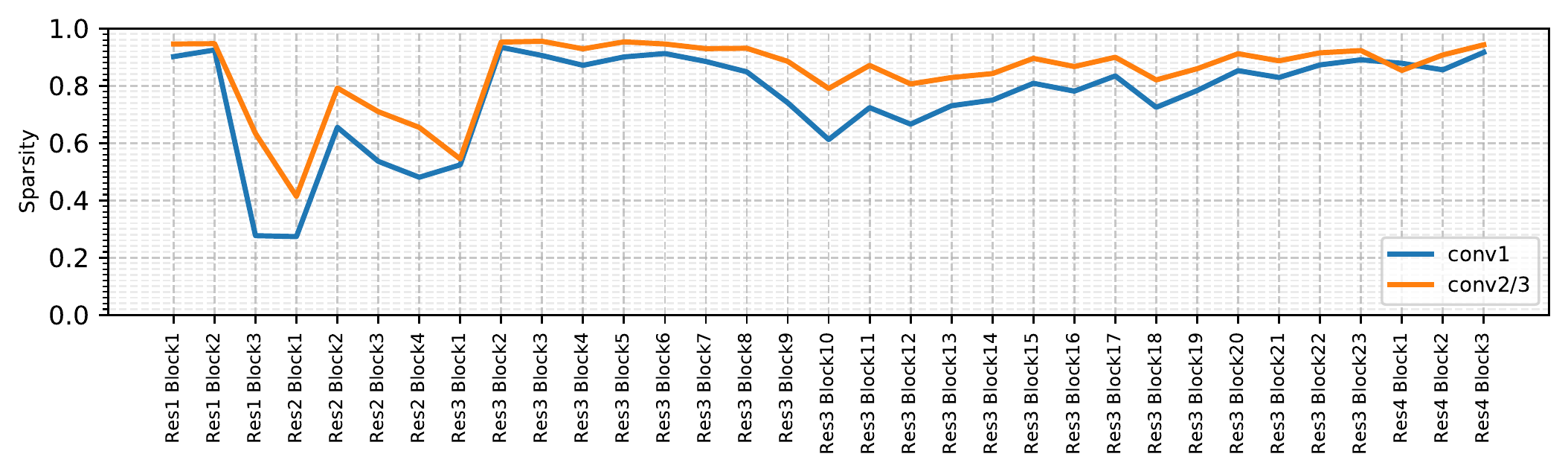} 
    }
  \caption{Sparsity of different residual blocks. The evaluated models are trained with different loss weights.}
  \label{fig:sparsity_diff_layers}
  \vspace{-0.5em}
\end{figure*}